\documentclass{article} %
\usepackage{hyperref}
\usepackage{cite}

\usepackage{geometry}
\usepackage{setspace}
\usepackage[square,sort,compress]{natbib}
\usepackage{times}
\usepackage{ragged2e}
\usepackage{float}
\usepackage{latexsym}
\usepackage{url}
\usepackage{multirow}
\usepackage[ruled,vlined]{algorithm2e}
\usepackage{algorithmic}
\usepackage[utf8]{inputenc} 
\usepackage[T1]{fontenc} 
\usepackage{hyperref} 
\usepackage{url} 
\usepackage{booktabs} 
\usepackage{amsfonts} 
\usepackage{nicefrac} 
\usepackage{microtype} 
\usepackage[utf8]{inputenc} 
\usepackage[T1]{fontenc} 
\usepackage{hyperref} 
\usepackage{url} 
\usepackage{booktabs} 
\usepackage{amsfonts} 
\usepackage{nicefrac} 
\usepackage{microtype} 
\usepackage{graphicx}
\usepackage{xcolor}
\usepackage[super]{nth}
\usepackage{gensymb}
\usepackage{subcaption}
\usepackage{amsthm}
\usepackage{mathrsfs}
\usepackage{amsmath}
\usepackage{amssymb}
\usepackage{mathtools}
\usepackage{wrapfig}
\usepackage{soul}
\usepackage{pifont}

\makeatletter
\def\BState{\State\hskip-\ALG@thistlm}
\makeatother

\usepackage{stackengine}

\makeatletter
\newcommand{\distas}[1]{\mathbin{\overset{#1}{\kern\z@\sim}}}%

\usepackage{algorithm2e}

\newcommand{\beqs}{\vspace{0mm}\begin{eqnarray}}
\newcommand{\eeqs}{\vspace{0mm}\end{eqnarray}}
\newcommand{\barr}{\begin{array}}
\newcommand{\earr}{\end{array}}

\usepackage{bbm}

\usepackage{amssymb}
\usepackage{pifont}

\newcommand{\ours}{{AutoML-GPT}}

\date{}

\title{
AutoML-GPT: Automatic Machine Learning with GPT
}

\author{Shujian Zhang  \qquad Chengyue Gong \qquad  Lemeng Wu \qquad Xingchao Liu \\ \qquad Mingyuan Zhou \\
The University of Texas at Austin\\
\texttt{\{szhang19, mzhou\}@utexas.edu} 
}

\begin{document}
\maketitle

\begin{abstract} 
AI tasks encompass a wide range of domains and fields. While numerous AI models have been designed for specific tasks and applications, they often require considerable human efforts in finding the right model architecture, optimization algorithm, and hyperparameters. Recent advances in large language models (LLMs) like ChatGPT show remarkable capabilities in various aspects of reasoning, comprehension, and interaction. 
Consequently, we propose developing task-oriented prompts and automatically utilizing LLMs to automate the training pipeline.
To implement this concept, we present the {\ours}, which employs GPT as the bridge to diverse AI models and dynamically trains models with optimized hyperparameters. 
{\ours} dynamically takes user requests from the model and data cards and composes the corresponding prompt paragraph. 
Ultimately, with this prompt paragraph, AutoML-GPT will automatically conduct the experiments from data processing to model architecture, hyperparameter tuning, and predicted training log. 
By leveraging {\ours}'s robust language capabilities and the available AI models, {\ours} can tackle numerous intricate AI tasks across various tasks and datasets. This approach achieves remarkable results in computer vision, natural language processing, and other challenging areas. Extensive experiments and ablation studies demonstrate that our method can be general, effective, and beneficial for many AI tasks. 

\end{abstract}

\section{Introduction}\label{sec:introduction}
Artificial intelligence (AI) has experienced significant advancements recently. 
Among these developments, ChatGPT \citep{openai2023gpt} has particularly stood out due to its ability to reason, comprehend, and interact \citep{wu2023visual}. The ability to execute new tasks based on instructions is a crucial step towards achieving artificial general intelligence, and
the remarkable capabilities of large language models (LLMs) have spurred numerous emerging research topics, such as in-context learning \citep{xie2021explanation, ram2023context}, chain-of-thought prompting \citep{wei2022chain, pilault2023interactive}, retrieve and read \citep{izacard2020leveraging, zhang2021knowing, zhang2022passage}, and GPT-based intelligent systems \citep{zheng2023can}. 
These areas aim to explore the vast potential of LLMs and present boundless opportunities for constructing sophisticated AI systems.

LLMs, such as GPT-4 \citep{brown2020language, openai2023gpt}, LLaMA \citep{touvron2023llama}, Flan-T5 \citep{chung2022scaling}, and PaLM \citep{chowdhery2022palm}, have demonstrated a deep comprehension of natural language and the capacity to produce coherent, contextually appropriate responses. 
This progress has opened up new potential applications for challenging tasks involving different domain data, such as image and text processing, as well as the incorporation of domain-specific knowledge. In this context, LLMs play a crucial role, as their capacity to comprehend and produce natural language allows AI to better understand and tackle a wide range of challenges. 

In this paper, we aim to develop an Automatic Machine Learning (AutoML) system called {\ours}, which utilizes LLMs to automatically train the models on datasets with user inputs and descriptions. 
The LLMs are employed as an automatic training system to establish connections with versatile models and process the inputs. 
We suggest using language as a universal interface and prompt for LLMs to interact with users. 
By incorporating both data and model descriptions into prompts, LLMs can manage AI models for data processing, model architecture design, and hyperparameter tuning. They can invoke these models as needed to tackle AI tasks and return the predicted training log.
However, incorporating multiple AI models into LLMs demands a substantial number of high-quality model descriptions. 
To overcome this challenge, we recommend tapping into both model card \citep{mitchell2019model} that provides well-defined model descriptions and data card \citep{gebru2021datasheets} for specific AI tasks. This approach would enable us to connect diverse models through a language-based interface, thus facilitating the solution of complex AI tasks. It can also enhance the transferability among models and datasets by capturing their similarity.

\begin{figure}[h]
\centering
\includegraphics[width=12.0cm]{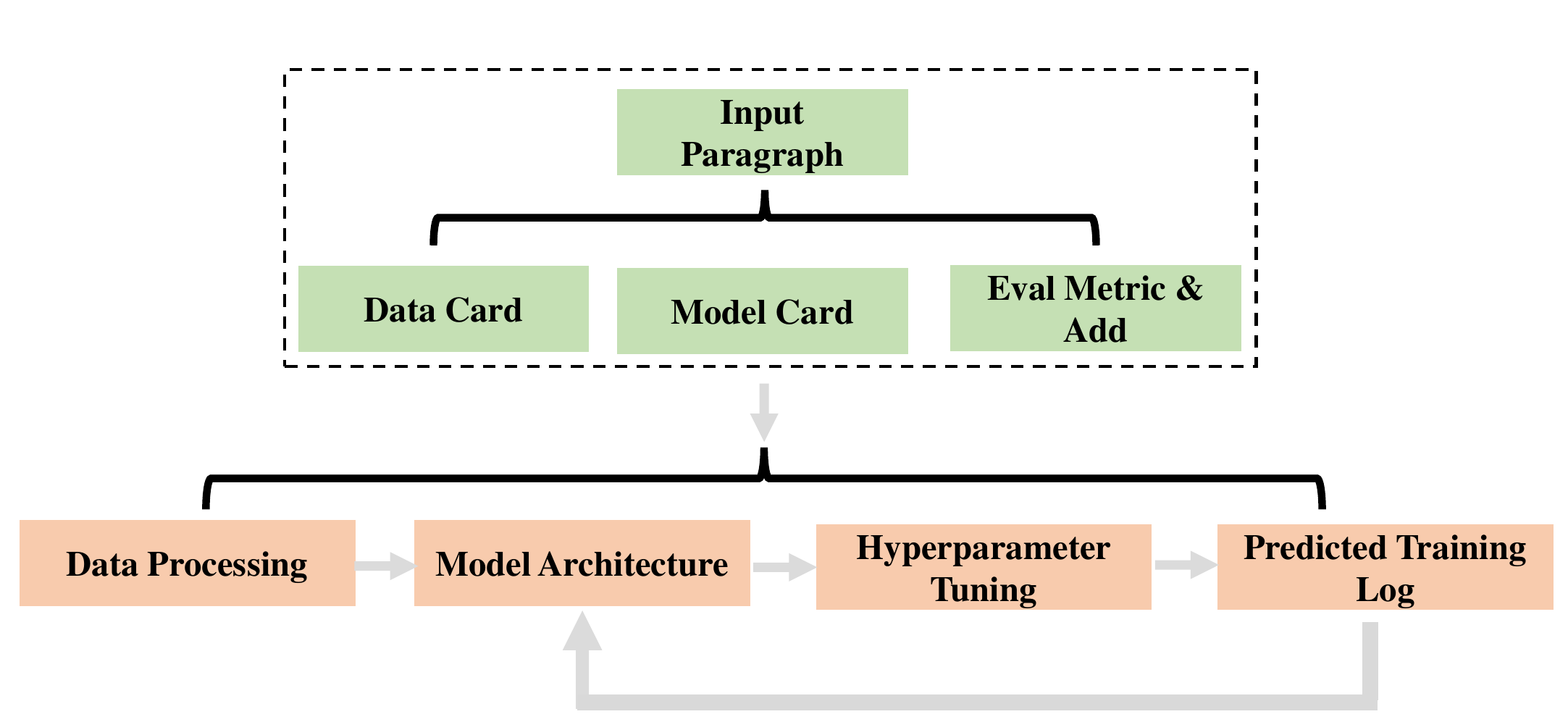}
\caption{Overview of {\ours}. Some notations are labeled along with corresponding components. `Eval Metrics \& Add' refers to the evaluation metrics and additional requests.
}
\label{fig:auto_ml_pipeline}
\end{figure}

{\ours} connects versatile machine learning models, training pipelines, and datasets to solve numerous complex AI tasks. More specifically, for each AI task we aim to solve, using
its corresponding description (such as model card and data card ), we  fuse the paragraph as the prompt into a pretrained LLMs (such as ChatGPT) to establish the AutoML pipeline. Afterward, in our system, LLMs 
perform the automatic training to
return the predicted training logs for the input questions of users. Based on these training logs, we can further interact with the LLM to solve requests (such as hyperparameter tuning) shown in Figure \ref{fig:auto_ml_pipeline}. Thus, the whole process of
{\ours} can be divided into four stages:
1) data processing, 2) model architecture design, 3) hyper-parameter tuning with the predicted training log, 4) human feedback on experimental data.

Benefiting from such a design, {\ours} in Figure \ref{fig:auto_ml_pipeline} is able to use external models and thus can handle multiple tasks on well-known benchmarks, and transfer the knowledge to unknown private dataset when only given metadata (data card). 
Furthermore, this pipeline
also allows {\ours} to continue absorbing the powers from task-specific experts, enabling
growable and scalable AI capabilities. 
In summary, our contributions are as follows:
\begin{itemize}
\setlength{\itemsep}{0pt}
\setlength{\parsep}{0pt}
\setlength{\parskip}{0pt}
    \item To complement the advantages of large language models and expert models, we propose {\ours}, 
    which acts as the system for data processing and model architecture design and automatically conducts the experiments for each specific task.
    \item By integrating the model card with model descriptions and the data card with data descriptions, 
    we provide a fixed-format prompt paragraph and build a training pipeline to tackle general AI tasks.
    \item 
    Extensive evaluations on multiple AI tasks across language, vision, and continual learning demonstrate the capability of {\ours} in auto training. It further demonstrates the effectiveness of providing the hyperparameter tuning for an unseen or new dataset.
\end{itemize}

\section{AutoML-GPT} \label{sec:method_section}
{\ours} is a collaborative system that relies on the data and model information to format the prompt input paragraph. 
The LLM serves as the controller, 
while numerous expert models as collaborative executors. The workflow of {\ours} consists
of four stages: data processing, model architecture design, hyper-parameter tuning, and training log generation. 
Specifically, we suggest a general recipe for {\ours}: 1) generate a fixed-format prompt paragraph with both the model card and data card, 2) build the training pipeline and process the user request on the selected dataset and model architectures, 3) generate the performance training log and tune the hyperparameters, and 4) tune the model with the auto-suggested hyperparameters. 

\subsection{Input Decomposition}
In the first stage of {\ours}, an LLM takes the input from the users. To boost the performance of the LLM and generate an effective prompt, we employ specific instructions for the input prompt. The instructions contain three parts described  below.

\begin{figure}[h]
\centering
\includegraphics[width=12.0cm]{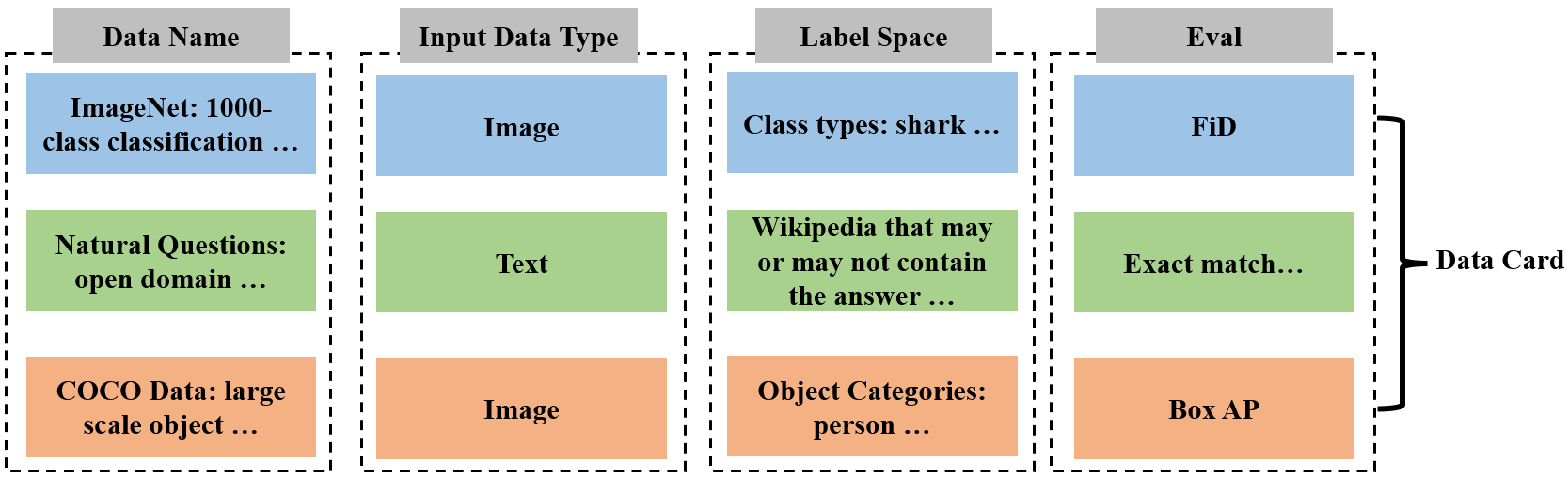}
\caption{The Data Card includes the data name, input data type, label space, and evaluation metric. Within the data card, the same color denotes information originating from a single dataset.
}
\label{fig:data_card}
\end{figure}

\paragraph{Data Card}
To clarify the intended use cases
of datasets and minimize their usage in contexts
for which they are not well suited, we utilize the data card that provides comprehensive documentation for this dataset. As shown in Figure \ref{fig:data_card}, the key components of the data card are comprised of the dataset name, input dataset type (\emph{e.g.}, image data or text data), label space (\emph{e.g.}, the class types or resolution), and default evaluation metrics.

\begin{figure}[h]
\centering
\includegraphics[width=12.0cm]{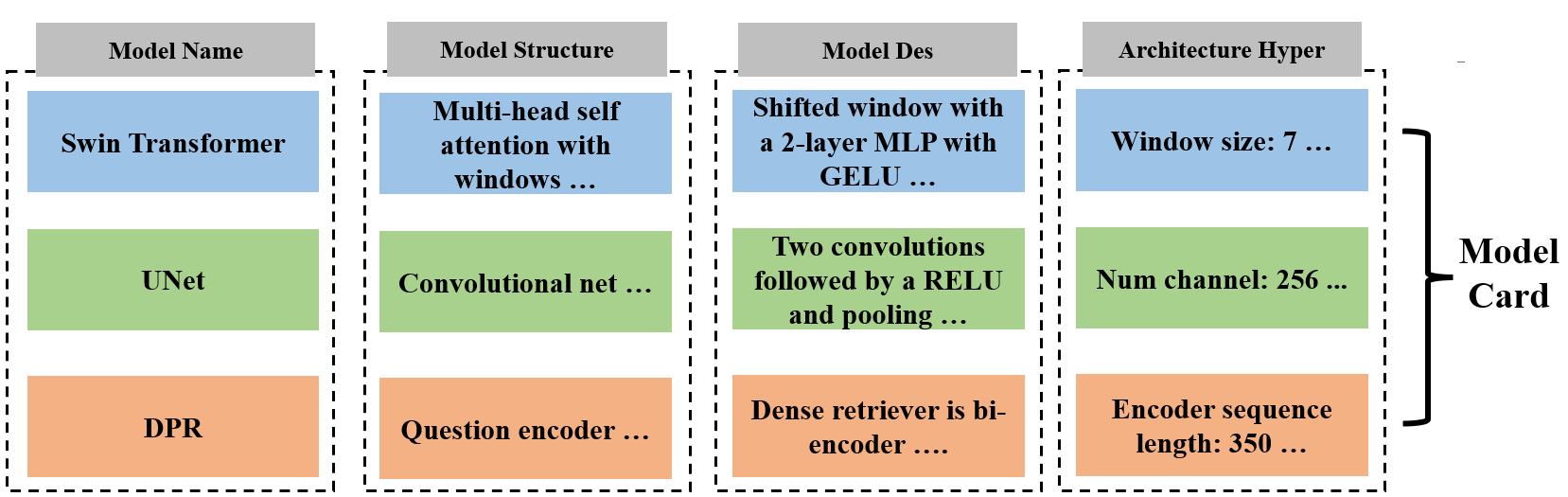}
\caption{The Model Card comprises model name, model structure, model descriptions, and architecture hyperparameters. In the model card, the same color represents information from a single model card.
}
\label{fig:model_card}
\end{figure}

\paragraph{Model Card}
The model cards in Figure \ref{fig:model_card}, complementary to the ``Data Card'' discussed earlier, serve as one of
the proposed paradigms that report details of
the model used to train and test the datasets. The model card consists of the model name, model structure (\emph{e.g.}, Swin transformer \citep{liu2021swin} with a UperNet \citep{xiao2018unified} head), model descriptions, and architecture hyperparameter. 
By providing this information, model cards inform the LLM about the machine learning systems used and the degree of flexibility the user would like to have on the model architecture. It would further create more inclusive outcomes with the LLM. 

\paragraph{Evaluation Metrics and Additional Requests}
In addition to the model cards and data cards, users can have the option to request more evaluation benchmarks, metrics, or any constraints. 
Except for the default evaluation metrics, we can add specific metrics or constraints according to the user's request when selecting the model architecture. 
For example, given a constraint ``the inference time smaller than 10 FPS,'' we then process the user requests under the evaluation metrics and constraints. 
Benefiting from this instruction and human feedback of these evaluation metrics and additional requests, the LLM can follow instructions better. {\ours} provides these task specifications
to the LLM as high-level instructions for analyzing the user’s requests accordingly.

\subsection{Data Processing}
Data processing is an integral step in machine learning as the quality of data 
and the derived useful information 
directly affect the ability of our model to learn. 
It is thus crucial that we process the data before feeding it into our model. 
For example, in computer vision, data processing refers to the set of techniques and methods used to prepare raw image data for analysis or machine learning algorithms. This can include image resizing, normalization, augmentation, and filtering.
Similarly, in Natural Language Processing (NLP) projects, data processing refers to transforming raw text data into a structured and clean format that machine learning algorithms can easily understand and process. Techniques such as tokenization, stopword removal, lowercasing, and removal of special characters and numbers are commonly used. Based on the provided data card and data descriptions, {\ours} provides specific process techniques depending on the project's requirements and the data's nature.

\subsection{Model Architecture}
Upon processing the list of tasks, {\ours} needs to match each task with a corresponding model, essentially selecting the suitable model for every task in the list. To achieve this, we first acquire model cards and descriptions of the models from the user inputs. Following that, we dynamically assign models to tasks using the in-context task-model assignment mechanism. 
This approach enables incremental model access and offers greater openness and flexibility by combining the providing model descriptions and a better understanding of the user requests.

Model architectures refer to detailed explanations of a machine learning model's design, structure, and components. These descriptions typically include the following elements: input and output layers, hidden layers, activation functions, loss functions, and model-specific components (such as attention mechanisms, convolutional layers, or recurrent layers).

\begin{figure*}[h]
\centering
\includegraphics[width=15.0cm]{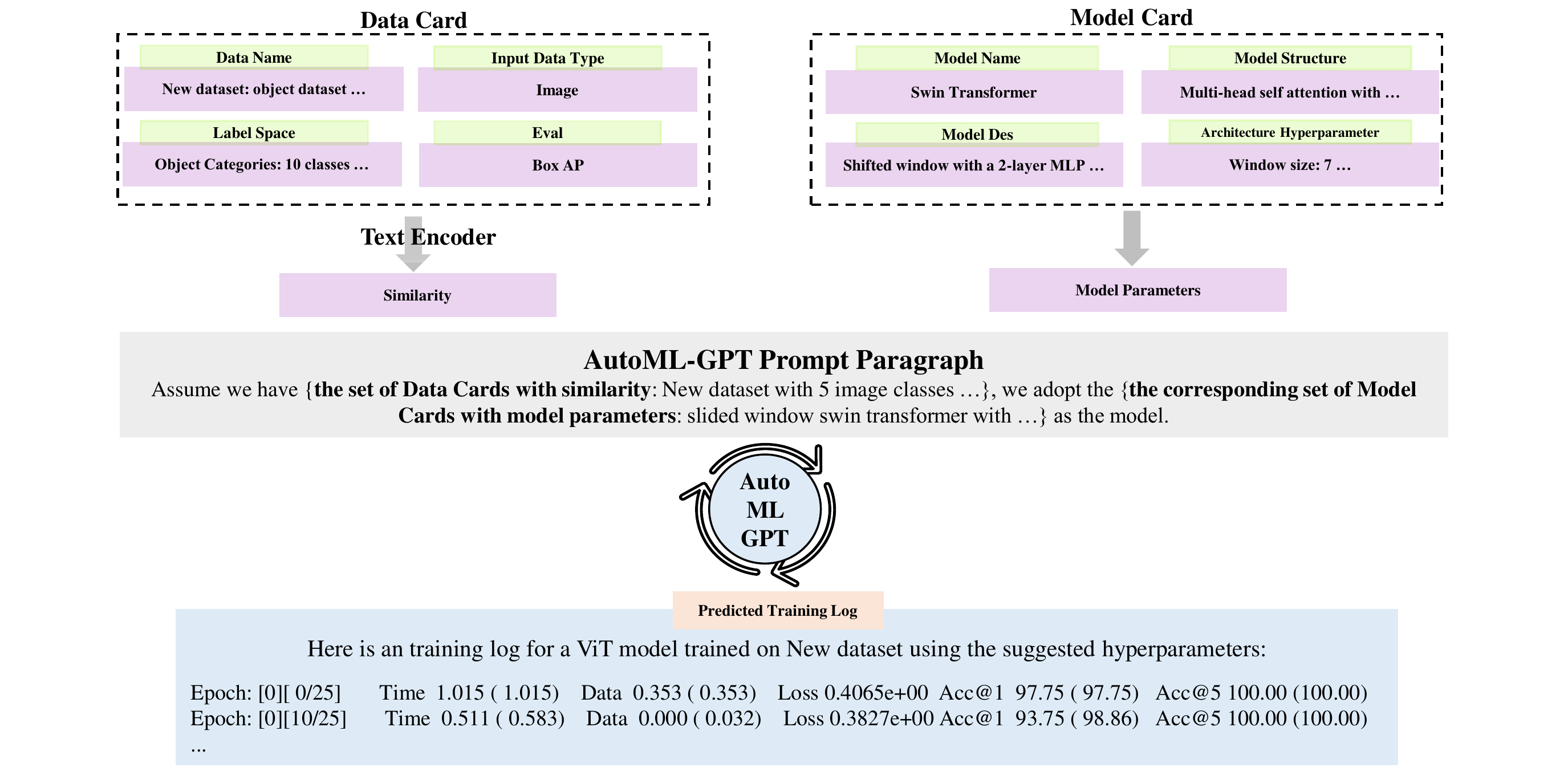}
\caption{Overview of {\ours} for the unseen dataset: the top block showcases data card and model information. 
We first log the training information for several datasets.
The data cards for these datasets are processed through a text encoder to obtain similarity scores, which are then combined with model parameters of corresponding trained models to form the AutoML-GPT prompt paragraph. The bottom block presents the predicted training log based on the recommended hyperparameter settings for the unseen dataset.
}
\label{fig:demo_unseen_dataset}
\end{figure*}
\subsection{Hyperparameter Tuning with Predicted Training Log}
To find the optimal set of hyperparameters that yield the best performance for a given model on a specific dataset, hyperparameter tuning is a crucial step in machine learning. Hyperparameters are configuration settings that are not learned during the training process but are predefined and control various aspects of the model's learning behavior. Examples of common hyperparameters include the learning rate, batch size, number of hidden layers, and number of neurons per layer.

In order to tune hyper-parameters without training on real machines, we predict the performance by generating a training log for a given hyper-parameter setting for the provided data card and model card. 
{\ours} will automatically conduct the training and return the training log. The training log of model performance on a dataset records various metrics and information collected during the training process. It helps in understanding the model's progress, identifying potential issues, and evaluating the effectiveness of the chosen architecture, hyperparameters, and optimization techniques. A typical training log includes the epoch numbers with training and validation metrics. By examining the training log, we can form a basic understanding of the model performance according to the user feedback.

\paragraph{Unseen Datasets}
The hyperparameter tuning for unseen private datasets could be even more challenging. 
Given the metadata of an unseen dataset, {\ours} can recommend a hyperparameter configuration that is likely to be effective for that dataset.
We rely on the data card to leverage the necessary text descriptions and identify the correlation between the unseen dataset and the existing ones.
Based on the correlation, we transfer the hyper-parameter settings from the existing datasets to the new unseen dataset.

To calculate the correlation, we use a text encoder to encode the data card.
Specifically, in the data card, it contains information such as class type, resolution, image size, and other relevant metadata.
We take the dataset scale, task description, label space, and input/output data type as the input to a text encoder (\emph{e.g.}, CLIP \citep{radford2021learning}) and describe the correlation between this unseen dataset and the existing datasets using the similarity score of the encoded latent representation.

\section{Experiments}
We assess the performance of our AutoML-GPT and implement it using ChatGPT (OpenAI's ``GPT-4'' version)\footnote{\url{https://platform.openai.com/}}. Various case studies are carried out to showcase the efficacy of our approach from multiple angles.

\begin{figure*}[h]
\centering
\includegraphics[width=15.0cm]{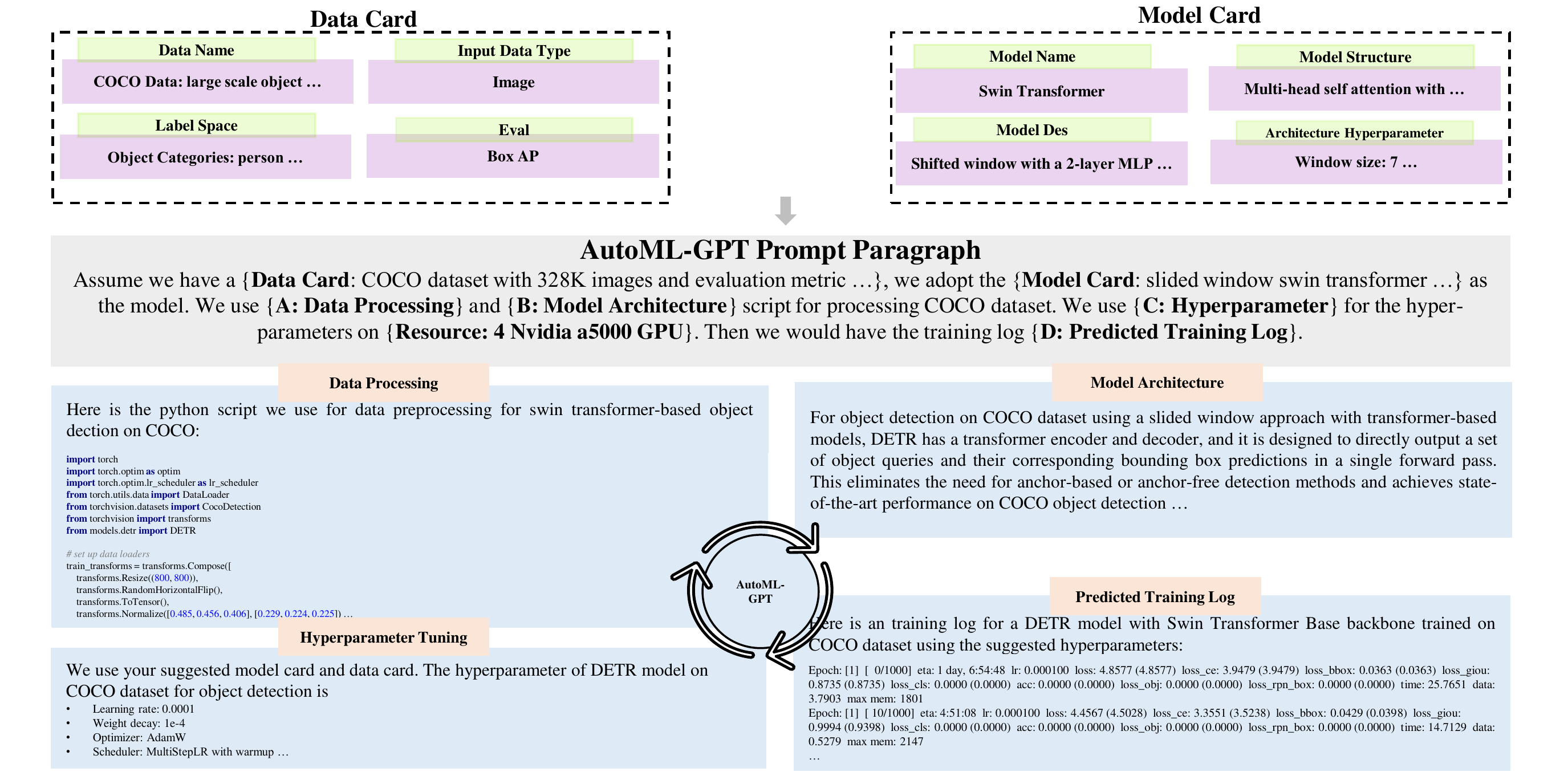}
\caption{Overview of {\ours} for object detection: The top block displays the data card and model card. The middle block showcases the AutoML-GPT prompt paragraph, derived from the data card and model card.
The bottom block outlines the four steps: data processing, model architecture, hyperparameter tuning, and predicted training log. We use the predicted training log to tune the hyperparameters before feedbacking the hyperparameters to the users. 
}
\label{fig:demo_coco}
\end{figure*}

\subsection{Unseen Dataset}\label{sec:unseen_dataset_section}
In Figure \ref{fig:demo_unseen_dataset}, we present the results of training on an unseen dataset using AutoML-GPT. 
To verify the performance in real cases, we construct a set of performance and hyper-parameters on already trained datasets, and some coming untrained datasets. We will predict hyperparameter configurations for these untrained datasets.
We make our test environment based on the classification setting described in \citet{vinyals2016matching}. We also follow the MiniImageNet \citep{vinyals2016matching} to subsample and split the training dataset \citep{deng2009imagenet} into 80\% and 20\% portions.
From the 80\% data, 
we construct the data cards and corresponding model cards (containing model best hyperparameters). We randomly select fifteen classes to create various subset datasets (\emph{e.g.}, dataset A, B, etc.), grid search the hyper-parameters, finetune the ViT base model \citep{dosovitskiy2020image} and log the best performance on these subset datasets. 
We then create a new dataset called ``New'' with ten image classes from the remaining 20\% data.

\begin{figure*}[h]
\centering
\includegraphics[width=15.0cm]{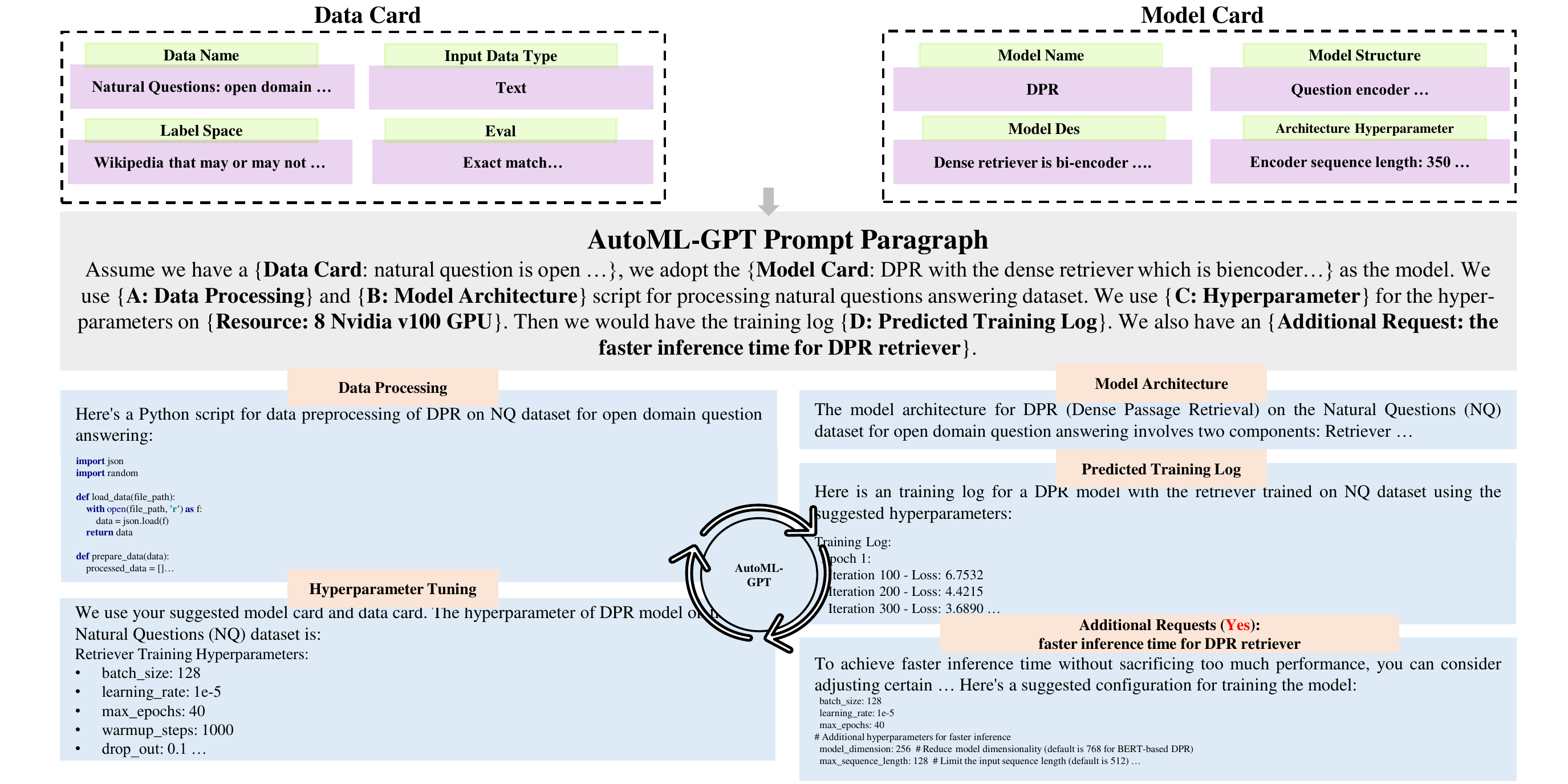}
\caption{Overview of {\ours} for question answering: The top block presents data card and model information, while the middle block highlights the AutoML-GPT prompt paragraph, derived from both data card and model card. The bottom block details the four steps: data processing, model architecture, hyperparameter tuning, and predicted training log. 
}
\label{fig:demo_nq}
\end{figure*}

\begin{figure*}[h]
\centering
\includegraphics[width=15.0cm]{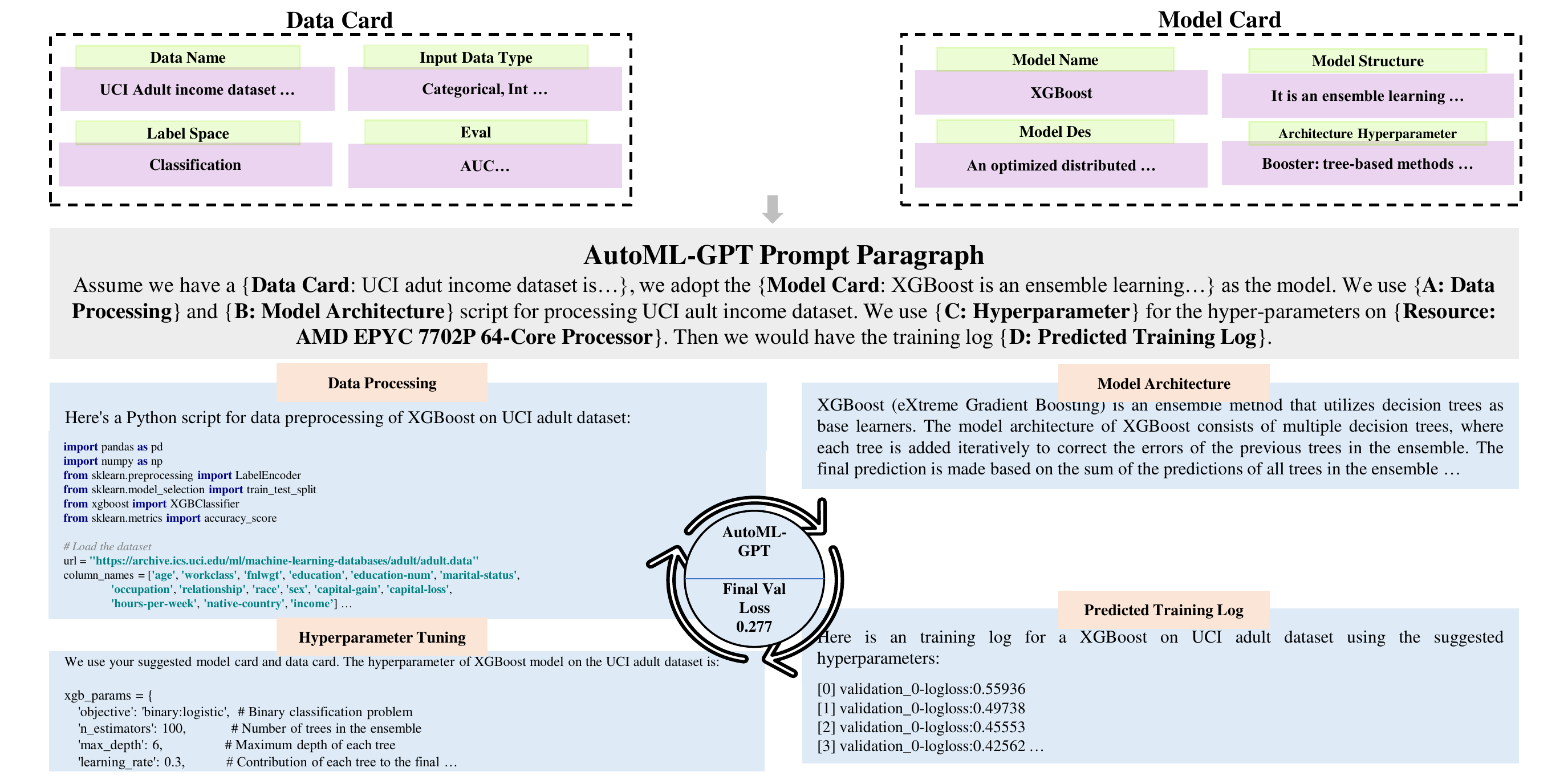}
\caption{Overview of {\ours} for classification: The top block displays data card and model information, and the middle block showcases the AutoML-GPT prompt paragraph, derived from both data card and model card. The bottom block outlines the four steps: data processing, model architecture, hyperparameter tuning, and predicted training log. Additionally, we include the final validation results, following the hyperparameter recommendations from {\ours} and training the model.
}
\label{fig:demo_xgboost}
\end{figure*}

To demonstrate the capabilities of our approach on unseen datasets, we utilize AutoML-GPT to recommend the best training configuration for the ``New'' dataset based on the provided data card and model card.
In our data card, we log the label space, $i.e.$, text descriptions for each class.
In practice, we incorporate a similarity score between two data cards by passing the text in the data card through a text encoder, \emph{e.g.}, the CLIP text encoder, and calculating the similarity.
Specifically, in Figure \ref{fig:demo_unseen_dataset}, we state that the ``New'' dataset has a 60\% label space similarity to dataset A and a 40\% label space similarity to dataset B. Using this information and the hyper-parameter settings in the data cards for dataset A and B, AutoML-GPT can recommend the appropriate hyperparameter settings for training on the ``New'' dataset.
In our experiments, we achieve 98\% accuracy for the Top 1 prediction, compared to 80\% Top 1 accuracy with average random-selected hyperparameters.
Moreover, we also initialize the model using the suggested hyperparameter settings from {\ours} without giving any additional datasets
With this configuration, we achieve 82\% Top 1 accuracy, which is better than the average randomly-selected hyperparameters but not as good as our recommended setting. 
It also suggests that ChatGPT can give good hyperparameter settings for a specific task ($e.g.$, image classification).
This demonstrates the effectiveness of our proposed auto-training approach in addressing machine learning problems, even with unseen or new datasets.
These findings highlight the potential of our auto-training method to enhance machine learning by providing accurate hyperparameter recommendations.

\subsection{Object Detection}\label{sec:coco_detection_section}

Figure~\ref{fig:demo_coco} presents our results on the COCO dataset \citep{lin2014microsoft} for object detection. \ding{192} The top block displays the data card for the COCO dataset and the model card for ImageNet, based on user input. The middle block demonstrates the AutoML-GPT Prompt Paragraph derived from the input decomposition. The information from the data card and model card is automatically incorporated into our prompt format. We report the results for data processing, model architecture design, hyperparameter tuning, and 
training log generation.
\ding{193} In data processing, AutoML-GPT generates a script for handling the input dataset. We also provide a Python script example in Figure \ref{fig:demo_coco}. For model architecture design, our pipeline generates a model composition for subsequent training. Once both the data and model are prepared, the detailed configurations are provided in the hyperparameter-tuning stage ($e.g.$,  learning rate: $10^{-4}$, weight decay: $10^{-4}$) and are further tuned with predicted training logs.
\ding{194} These results further validate that our method can serve as an effective pipeline for flexibly adapting LLMs to downstream tasks. Our approach, which employs data and model cards to derive the AutoML-GPT prompt paragraph, can also be considered as a complementary module for works focused on enhancing LLM prompt components.

\subsection{Question Answering}\label{sec:open_qa_section}
We present the experimental results on the Natural Questions Open dataset \citep{kwiatkowski2019natural} in Figure \ref{fig:demo_nq}. We utilize Dense Passage Retrieval (DPR) \citep{karpukhin2020dense}. \ding{172} For the data card, users input the data name, input data type, label space, and evaluation metrics. \ding{173} For the model card, it includes model name, model structure, model descriptions, and architecture hyperparameters. 
\ding{174} With the generated AutoML-GPT prompt paragraph, {\ours} carries out data processing, model architecture creation, hyperparameter tuning, and generates a predicted training log.
As seen in the ``Hyperparameter Tuning,'' the hyperparameters generated by {\ours} and those provided by DPR align closely, $e.g.$, the learning rate is $10^{-5}$ and max epochs is 40.
\ding{175} Once the predicted training log is available, we showcase a scenario where the user can ask {\ours} for different evaluation metrics or model architectures based on their requirements, as illustrated in Figure \ref{fig:demo_nq} ``Additional requests: fast inference time for DPR retriever.'' 
As seen in the returned response in Figure \ref{fig:demo_nq}, AutoML-GPT also offers hints such as ``without sacrificing too much performance.'' 
{\ours} further tunes the hyper-parameters based on these requests and predicted logs.
Our method demonstrates the powerful ability to automatically conduct experiments and perform interactive hyperparameter tuning. It further confirms that our approach works well for various datasets and can generalize across different input types and domains.

\subsection{Classification}\label{sec:classification_section}
We also evaluate {\ours} on the UCI Adult dataset \citep{Dua:2019} using XGBoost. As before, we supply the data card and model card to generate the input prompt paragraph. The same training pipeline is applied here, as shown in Figure \ref{fig:demo_xgboost}. 
We also adhere to the hyperparameter settings suggested by {\ours} and train the XGBoost model. This training results in a final validation loss of 0.277 with 85.92\% accuracy.
Despite the different inputs and tasks, our proposed AutoML-GPT consistently delivers strong performance in classification. This further demonstrates that {\ours} can be employed for a wide range of machine learning problems across various tasks.

\section{Related Work} \label{sec:related_work_section}

\paragraph{Advanced Large Language Model}
LLMs have exhibited robustness and generalizability through zero-shot and few-shot learning by having parameter sizes exceeding one hundred billion. Notable examples of LLMs include Megatron-turing NLG \citep{smith2022using} with 530 billion
parameters, Gopher \citep{rae2021scaling} with 280 billion parameters, and PaLM \citep{chowdhery2022palm} with 540 billion
parameters. The scaling of LLM has unlocked new emergent abilities previously unobserved under smaller models \citep{wei2022emergent}. 
These LLMs have demonstrated the superiority of LLMs for zero-shot learning.
Among existing LLMs, ChatGPT has unique characteristics. It has the ability to interact
with users in a conversation-like manner, while
retaining its accumulated knowledge and generalization ability gained from pre-training. 
Going a step further, we explore the zero-shot learning capability of ChatGPT on different tasks beyond dialogue in this work.

\paragraph{Chain of Thought}
Chain-of-thought (CoT) prompting induces LLMs to generate intermediate reasoning steps before answering \citep{wei2022chain}. 
There are two lines of research focusing on the current CoT prompting. One line is exploring the manually designed CoT. In the manually designed CoT, LLMs adapt the manually designed features and demonstration for the reasoning process \citep{wei2022chain}. \citet{wang2022self} proposes
a new decoding strategy, self-consistency, to replace the naive greedy decoding used in chain-of-thought prompting. 
Recently, Interactive-Chain-Prompting \citep{pilault2023interactive} is introduced to resolve the ambiguity for crosslingual conditional generation.
Another line is conducting research on the zero-shot setting, where STaR \citep{zelikman2022star} is introduced for the self-generation and helps the model to self-improve, and Automatic Reasoning and Tool-use
(ART) \citep{paranjape2023art} is a framework that uses frozen LLMs to automatically generate intermediate reasoning steps as a program. 

\paragraph{GPT-based Systems}
GPT \citep{brown2020language} has shown promising
performance improvements. A recent line of research has focused on integrating the GPT model into AI systems. HuggingGPT \citep{shen2023hugginggpt} is built with the HuggingFace transformers library and utilizes the GPT as the interaction agent. VisualGPT \citep{wu2023visual} incorporates different
Visual Foundation Models to enable the user to interact
with ChatGPT.
OpenAGI \citep{ge2023openagi}, an open-source AGI research platform, is designed to offer
complex, multi-step tasks and accompany by task-specific datasets.
Similarly, we also integrate the GPT into our AutoML pipeline. 
There is also another GPT based system that can incorporate extra information from  search engines, $e.g.$, AutoGPT \footnote{\url{https://github.com/Significant-Gravitas/Auto-GPT}}.
{\ours} rethinks the impact of ChatGPT from the auto training perspective. We focus on building the training pipeline and establishing an AutoML system from the start to end.

\section{Conclusion}
Our work demonstrates the benefits of building AutoML systems upon GPT. The proposed method can automatically conduct machine learning experiments. This automatic learning dramatically improves training efficiency and enhances the model's performance.
We demonstrate use cases across computer vision, natural questions answering, and classification benchmarks. We further conduct a detailed use case with the unseen datasets and additional interactions between the user and AutoML-GPT.
To summarize, the proposed {\ours} is effective and general, with the potential to create a natural language interface for tuning machine learning models for various tasks.
In the future, we will 1) automatically generate the model and data cards for well-known benchmarks and make them a part of our system, and 2) extract task-aware sub-networks from large pretrained models with the help of ChatGPT.

\label{sec:limitations}

\bibliography{naacl2021}

\begin{thebibliography}{34}
\expandafter\ifx\csname natexlab\endcsname\relax\def\natexlab#1{#1}\fi

\bibitem[{Brown et~al.(2020)Brown, Mann, Ryder, Subbiah, Kaplan, Dhariwal,
  Neelakantan, Shyam, Sastry, Askell et~al.}]{brown2020language}
Tom Brown, Benjamin Mann, Nick Ryder, Melanie Subbiah, Jared~D Kaplan, Prafulla
  Dhariwal, Arvind Neelakantan, Pranav Shyam, Girish Sastry, Amanda Askell,
  et~al. 2020.
\newblock Language models are few-shot learners.
\newblock \emph{Advances in neural information processing systems},
  33:1877--1901.

\bibitem[{Chowdhery et~al.(2022)Chowdhery, Narang, Devlin, Bosma, Mishra,
  Roberts, Barham, Chung, Sutton, Gehrmann et~al.}]{chowdhery2022palm}
Aakanksha Chowdhery, Sharan Narang, Jacob Devlin, Maarten Bosma, Gaurav Mishra,
  Adam Roberts, Paul Barham, Hyung~Won Chung, Charles Sutton, Sebastian
  Gehrmann, et~al. 2022.
\newblock Palm: Scaling language modeling with pathways.
\newblock \emph{arXiv preprint arXiv:2204.02311}.

\bibitem[{Chung et~al.(2022)Chung, Hou, Longpre, Zoph, Tay, Fedus, Li, Wang,
  Dehghani, Brahma et~al.}]{chung2022scaling}
Hyung~Won Chung, Le~Hou, Shayne Longpre, Barret Zoph, Yi~Tay, William Fedus,
  Eric Li, Xuezhi Wang, Mostafa Dehghani, Siddhartha Brahma, et~al. 2022.
\newblock Scaling instruction-finetuned language models.
\newblock \emph{arXiv preprint arXiv:2210.11416}.

\bibitem[{Deng et~al.(2009)Deng, Dong, Socher, Li, Li, and
  Fei-Fei}]{deng2009imagenet}
Jia Deng, Wei Dong, Richard Socher, Li-Jia Li, Kai Li, and Li~Fei-Fei. 2009.
\newblock Imagenet: A large-scale hierarchical image database.
\newblock In \emph{2009 IEEE conference on computer vision and pattern
  recognition}, pages 248--255. Ieee.

\bibitem[{Dosovitskiy et~al.(2020)Dosovitskiy, Beyer, Kolesnikov, Weissenborn,
  Zhai, Unterthiner, Dehghani, Minderer, Heigold, Gelly
  et~al.}]{dosovitskiy2020image}
Alexey Dosovitskiy, Lucas Beyer, Alexander Kolesnikov, Dirk Weissenborn,
  Xiaohua Zhai, Thomas Unterthiner, Mostafa Dehghani, Matthias Minderer, Georg
  Heigold, Sylvain Gelly, et~al. 2020.
\newblock An image is worth 16x16 words: Transformers for image recognition at
  scale.
\newblock \emph{arXiv preprint arXiv:2010.11929}.

\bibitem[{Dua and Graff(2017)}]{Dua:2019}
Dheeru Dua and Casey Graff. 2017.
\newblock \href {http://archive.ics.uci.edu/ml} {{UCI} machine learning
  repository}.

\bibitem[{Ge et~al.(2023)Ge, Hua, Ji, Tan, Xu, and Zhang}]{ge2023openagi}
Yingqiang Ge, Wenyue Hua, Jianchao Ji, Juntao Tan, Shuyuan Xu, and Yongfeng
  Zhang. 2023.
\newblock Openagi: When llm meets domain experts.
\newblock \emph{arXiv preprint arXiv:2304.04370}.

\bibitem[{Gebru et~al.(2021)Gebru, Morgenstern, Vecchione, Vaughan, Wallach,
  Iii, and Crawford}]{gebru2021datasheets}
Timnit Gebru, Jamie Morgenstern, Briana Vecchione, Jennifer~Wortman Vaughan,
  Hanna Wallach, Hal~Daum{\'e} Iii, and Kate Crawford. 2021.
\newblock Datasheets for datasets.
\newblock \emph{Communications of the ACM}, 64(12):86--92.

\bibitem[{Izacard and Grave(2020)}]{izacard2020leveraging}
Gautier Izacard and Edouard Grave. 2020.
\newblock Leveraging passage retrieval with generative models for open domain
  question answering.
\newblock \emph{arXiv preprint arXiv:2007.01282}.

\bibitem[{Karpukhin et~al.(2020)Karpukhin, O{\u{g}}uz, Min, Wu, Edunov, Chen,
  and Yih}]{karpukhin2020dense}
Vladimir Karpukhin, Barlas O{\u{g}}uz, Sewon Min, Ledell Wu, Sergey Edunov,
  Danqi Chen, and Wen-tau Yih. 2020.
\newblock Dense passage retrieval for open-domain question answering.
\newblock \emph{Empirical Methods in Natural Language Processing (EMNLP)}.

\bibitem[{Kwiatkowski et~al.(2019)Kwiatkowski, Palomaki, Redfield, Collins,
  Parikh, Alberti, Epstein, Polosukhin, Kelcey, Devlin, Lee, Toutanova, Jones,
  Chang, Dai, Uszkoreit, Le, and Petrov}]{kwiatkowski2019natural}
Tom Kwiatkowski, Jennimaria Palomaki, Olivia Redfield, Michael Collins, Ankur
  Parikh, Chris Alberti, Danielle Epstein, Illia Polosukhin, Matthew Kelcey,
  Jacob Devlin, Kenton Lee, Kristina~N. Toutanova, Llion Jones, Ming-Wei Chang,
  Andrew Dai, Jakob Uszkoreit, Quoc Le, and Slav Petrov. 2019.
\newblock Natural {Q}uestions: a benchmark for question answering research.
\newblock \emph{TACL}.

\bibitem[{Lin et~al.(2014)Lin, Maire, Belongie, Hays, Perona, Ramanan,
  Doll{\'a}r, and Zitnick}]{lin2014microsoft}
Tsung-Yi Lin, Michael Maire, Serge Belongie, James Hays, Pietro Perona, Deva
  Ramanan, Piotr Doll{\'a}r, and C~Lawrence Zitnick. 2014.
\newblock Microsoft coco: Common objects in context.
\newblock In \emph{Computer Vision--ECCV 2014: 13th European Conference,
  Zurich, Switzerland, September 6-12, 2014, Proceedings, Part V 13}, pages
  740--755. Springer.

\bibitem[{Liu et~al.(2021)Liu, Lin, Cao, Hu, Wei, Zhang, Lin, and
  Guo}]{liu2021swin}
Ze~Liu, Yutong Lin, Yue Cao, Han Hu, Yixuan Wei, Zheng Zhang, Stephen Lin, and
  Baining Guo. 2021.
\newblock Swin transformer: Hierarchical vision transformer using shifted
  windows.
\newblock In \emph{Proceedings of the IEEE/CVF international conference on
  computer vision}, pages 10012--10022.

\bibitem[{Mitchell et~al.(2019)Mitchell, Wu, Zaldivar, Barnes, Vasserman,
  Hutchinson, Spitzer, Raji, and Gebru}]{mitchell2019model}
Margaret Mitchell, Simone Wu, Andrew Zaldivar, Parker Barnes, Lucy Vasserman,
  Ben Hutchinson, Elena Spitzer, Inioluwa~Deborah Raji, and Timnit Gebru. 2019.
\newblock Model cards for model reporting.
\newblock In \emph{Proceedings of the conference on fairness, accountability,
  and transparency}, pages 220--229.

\bibitem[{OpenAI(2023)}]{openai2023gpt}
OpenAI. 2023.
\newblock Gpt-4 technical report.
\newblock \emph{arXiv}.

\bibitem[{Paranjape et~al.(2023)Paranjape, Lundberg, Singh, Hajishirzi,
  Zettlemoyer, and Ribeiro}]{paranjape2023art}
Bhargavi Paranjape, Scott Lundberg, Sameer Singh, Hannaneh Hajishirzi, Luke
  Zettlemoyer, and Marco~Tulio Ribeiro. 2023.
\newblock Art: Automatic multi-step reasoning and tool-use for large language
  models.
\newblock \emph{arXiv preprint arXiv:2303.09014}.

\bibitem[{Pilault et~al.(2023)Pilault, Garcia, Bra{\v{z}}inskas, and
  Firat}]{pilault2023interactive}
Jonathan Pilault, Xavier Garcia, Arthur Bra{\v{z}}inskas, and Orhan Firat.
  2023.
\newblock Interactive-chain-prompting: Ambiguity resolution for crosslingual
  conditional generation with interaction.
\newblock \emph{arXiv preprint arXiv:2301.10309}.

\bibitem[{Radford et~al.(2021)Radford, Kim, Hallacy, Ramesh, Goh, Agarwal,
  Sastry, Askell, Mishkin, Clark et~al.}]{radford2021learning}
Alec Radford, Jong~Wook Kim, Chris Hallacy, Aditya Ramesh, Gabriel Goh,
  Sandhini Agarwal, Girish Sastry, Amanda Askell, Pamela Mishkin, Jack Clark,
  et~al. 2021.
\newblock Learning transferable visual models from natural language
  supervision.
\newblock In \emph{International conference on machine learning}, pages
  8748--8763. PMLR.

\bibitem[{Rae et~al.(2021)Rae, Borgeaud, Cai, Millican, Hoffmann, Song,
  Aslanides, Henderson, Ring, Young et~al.}]{rae2021scaling}
Jack~W Rae, Sebastian Borgeaud, Trevor Cai, Katie Millican, Jordan Hoffmann,
  Francis Song, John Aslanides, Sarah Henderson, Roman Ring, Susannah Young,
  et~al. 2021.
\newblock Scaling language models: Methods, analysis \& insights from training
  gopher.
\newblock \emph{arXiv preprint arXiv:2112.11446}.

\bibitem[{Ram et~al.(2023)Ram, Levine, Dalmedigos, Muhlgay, Shashua,
  Leyton-Brown, and Shoham}]{ram2023context}
Ori Ram, Yoav Levine, Itay Dalmedigos, Dor Muhlgay, Amnon Shashua, Kevin
  Leyton-Brown, and Yoav Shoham. 2023.
\newblock In-context retrieval-augmented language models.
\newblock \emph{arXiv preprint arXiv:2302.00083}.

\bibitem[{Shen et~al.(2023)Shen, Song, Tan, Li, Lu, and
  Zhuang}]{shen2023hugginggpt}
Yongliang Shen, Kaitao Song, Xu~Tan, Dongsheng Li, Weiming Lu, and Yueting
  Zhuang. 2023.
\newblock Hugginggpt: Solving ai tasks with chatgpt and its friends in
  huggingface.
\newblock \emph{arXiv preprint arXiv:2303.17580}.

\bibitem[{Smith et~al.(2022)Smith, Patwary, Norick, LeGresley, Rajbhandari,
  Casper, Liu, Prabhumoye, Zerveas, Korthikanti et~al.}]{smith2022using}
Shaden Smith, Mostofa Patwary, Brandon Norick, Patrick LeGresley, Samyam
  Rajbhandari, Jared Casper, Zhun Liu, Shrimai Prabhumoye, George Zerveas,
  Vijay Korthikanti, et~al. 2022.
\newblock Using deepspeed and megatron to train megatron-turing nlg 530b, a
  large-scale generative language model.
\newblock \emph{arXiv preprint arXiv:2201.11990}.

\bibitem[{Touvron et~al.(2023)Touvron, Lavril, Izacard, Martinet, Lachaux,
  Lacroix, Rozi{\`e}re, Goyal, Hambro, Azhar et~al.}]{touvron2023llama}
Hugo Touvron, Thibaut Lavril, Gautier Izacard, Xavier Martinet, Marie-Anne
  Lachaux, Timoth{\'e}e Lacroix, Baptiste Rozi{\`e}re, Naman Goyal, Eric
  Hambro, Faisal Azhar, et~al. 2023.
\newblock Llama: Open and efficient foundation language models.
\newblock \emph{arXiv preprint arXiv:2302.13971}.

\bibitem[{Vinyals et~al.(2016)Vinyals, Blundell, Lillicrap, Wierstra
  et~al.}]{vinyals2016matching}
Oriol Vinyals, Charles Blundell, Timothy Lillicrap, Daan Wierstra, et~al. 2016.
\newblock Matching networks for one shot learning.
\newblock \emph{Advances in neural information processing systems}, 29.

\bibitem[{Wang et~al.(2022)Wang, Wei, Schuurmans, Le, Chi, and
  Zhou}]{wang2022self}
Xuezhi Wang, Jason Wei, Dale Schuurmans, Quoc Le, Ed~Chi, and Denny Zhou. 2022.
\newblock Self-consistency improves chain of thought reasoning in language
  models.
\newblock \emph{arXiv preprint arXiv:2203.11171}.

\bibitem[{Wei et~al.(2022{\natexlab{a}})Wei, Tay, Bommasani, Raffel, Zoph,
  Borgeaud, Yogatama, Bosma, Zhou, Metzler et~al.}]{wei2022emergent}
Jason Wei, Yi~Tay, Rishi Bommasani, Colin Raffel, Barret Zoph, Sebastian
  Borgeaud, Dani Yogatama, Maarten Bosma, Denny Zhou, Donald Metzler, et~al.
  2022{\natexlab{a}}.
\newblock Emergent abilities of large language models.
\newblock \emph{arXiv preprint arXiv:2206.07682}.

\bibitem[{Wei et~al.(2022{\natexlab{b}})Wei, Wang, Schuurmans, Bosma, Chi, Le,
  and Zhou}]{wei2022chain}
Jason Wei, Xuezhi Wang, Dale Schuurmans, Maarten Bosma, Ed~Chi, Quoc Le, and
  Denny Zhou. 2022{\natexlab{b}}.
\newblock Chain of thought prompting elicits reasoning in large language
  models.
\newblock \emph{arXiv preprint arXiv:2201.11903}.

\bibitem[{Wu et~al.(2023)Wu, Yin, Qi, Wang, Tang, and Duan}]{wu2023visual}
Chenfei Wu, Shengming Yin, Weizhen Qi, Xiaodong Wang, Zecheng Tang, and Nan
  Duan. 2023.
\newblock Visual chatgpt: Talking, drawing and editing with visual foundation
  models.
\newblock \emph{arXiv preprint arXiv:2303.04671}.

\bibitem[{Xiao et~al.(2018)Xiao, Liu, Zhou, Jiang, and Sun}]{xiao2018unified}
Tete Xiao, Yingcheng Liu, Bolei Zhou, Yuning Jiang, and Jian Sun. 2018.
\newblock Unified perceptual parsing for scene understanding.
\newblock In \emph{Proceedings of the European conference on computer vision
  (ECCV)}, pages 418--434.

\bibitem[{Xie et~al.(2021)Xie, Raghunathan, Liang, and Ma}]{xie2021explanation}
Sang~Michael Xie, Aditi Raghunathan, Percy Liang, and Tengyu Ma. 2021.
\newblock An explanation of in-context learning as implicit bayesian inference.
\newblock \emph{arXiv preprint arXiv:2111.02080}.

\bibitem[{Zelikman et~al.(2022)Zelikman, Wu, Mu, and
  Goodman}]{zelikman2022star}
Eric Zelikman, Yuhuai Wu, Jesse Mu, and Noah Goodman. 2022.
\newblock Star: Bootstrapping reasoning with reasoning.
\newblock \emph{Advances in Neural Information Processing Systems},
  35:15476--15488.

\bibitem[{Zhang et~al.(2021)Zhang, Gong, and Choi}]{zhang2021knowing}
Shujian Zhang, Chengyue Gong, and Eunsol Choi. 2021.
\newblock Knowing more about questions can help: Improving calibration in
  question answering.
\newblock \emph{arXiv preprint arXiv:2106.01494}.

\bibitem[{Zhang et~al.(2022)Zhang, Gong, and Liu}]{zhang2022passage}
Shujian Zhang, Chengyue Gong, and Xingchao Liu. 2022.
\newblock Passage-mask: A learnable regularization strategy for
  retriever-reader models.
\newblock \emph{arXiv preprint arXiv:2211.00915}.

\bibitem[{Zheng et~al.(2023)Zheng, Su, You, Wang, Qian, Xu, and
  Albanie}]{zheng2023can}
Mingkai Zheng, Xiu Su, Shan You, Fei Wang, Chen Qian, Chang Xu, and Samuel
  Albanie. 2023.
\newblock Can gpt-4 perform neural architecture search?
\newblock \emph{arXiv preprint arXiv:2304.10970}.

\end{thebibliography}
\bibliographystyle{acl_natbib}
\clearpage

\clearpage
\appendix



\end{document}